\documentclass{article}

\usepackage{microtype}
\usepackage{graphicx}
\usepackage{subfigure}
\usepackage{booktabs} 
\usepackage{amsmath}
\usepackage{amssymb}

\usepackage{float}

\usepackage{hyperref}

\newcommand{\corrglass}{\Delta}
\newcommand{\tw}{t_\mathrm{w}}

\usepackage[accepted]{icml2018}

\icmltitlerunning{Comparing Dynamics: Deep Neural Networks versus Glassy Systems}

\begin{document}

\twocolumn[
\icmltitle{Comparing Dynamics: Deep Neural Networks versus Glassy Systems}




\begin{icmlauthorlist}
\icmlauthor{Marco Baity-Jesi}{col} 
\icmlauthor{Levent Sagun}{cea,epfl} 
\icmlauthor{Mario Geiger}{epfl} 
\icmlauthor{Stefano Spigler}{epfl,cea} 
\icmlauthor{G\'erard Ben Arous}{courant} 
\icmlauthor{Chiara Cammarota}{kc} 
\icmlauthor{Yann LeCun}{courant,nyu,fb}  
\icmlauthor{Matthieu Wyart}{epfl} 
\icmlauthor{Giulio Biroli}{cea,ens} 
\end{icmlauthorlist}

\icmlaffiliation{cea}{Institut de Physique Th\'eorique, Universit\'e Paris Saclay, CEA, CNRS, F-91191 Gif-sur-Yvette, France}
\icmlaffiliation{col}{Department of Chemistry, Columbia University, New York, NY 10027, USA}
\icmlaffiliation{epfl}{EPFL, Lausanne, Switzerland}
\icmlaffiliation{kc}{King’s College London, Department of Mathematics, Strand, London WC2R 2LS, United Kingdom}
\icmlaffiliation{ens}{Laboratoire de Physique Statistique, \'Ecole Normale Sup\'erieure, CNRS, PSL Research University, Sorbonne Universit\'es, 75005 Paris, France}
\icmlaffiliation{courant}{Courant Institute of Mathematical Sciences, New York University, New York, USA}
\icmlaffiliation{nyu}{Center for Data Science, New York University, New York, USA}
\icmlaffiliation{fb}{Facebook AI Research, Facebook Inc., New York, USA}

\icmlcorrespondingauthor{Marco Baity-Jesi}{mb4399@columbia.edu}

\icmlkeywords{Machine Learning, ICML, Deep Learning, Dynamics, Slow Dynamics, Energy Landscape, Aging, Glass, Spin Glass}

\vskip 0.3in
]



\printAffiliationsAndNotice{} 

\begin{abstract}

We analyze numerically the training dynamics of deep neural networks (DNN) by using methods developed in statistical physics of glassy systems. The two main issues we address are (1) the complexity of the loss landscape and of the dynamics within it, and (2) to what extent DNNs share similarities with glassy systems. Our findings, obtained for different architectures and datasets, suggest that during the training process the dynamics slows down because of an increasingly large number of flat directions. 
At large times, when the loss is approaching zero, the system diffuses at the bottom of the landscape. Despite some similarities with the dynamics of mean-field glassy systems, in particular, the absence of barrier crossing, we find distinctive dynamical behaviors in the two cases, showing that the statistical properties of the corresponding loss and energy landscapes are \textit{different}. In contrast, when the network is under-parametrized we observe a typical glassy behavior, thus suggesting the existence of different phases depending on whether the network is under-parametrized or over-parametrized.

\end{abstract}

\section{Introduction}
\label{sec:intro}

The training process of a deep neural network (DNN) shares very strong similarities with the physical dynamics of disordered systems: the loss function plays the role of the energy, the weights are the degrees of freedom, and the dataset corresponds to the parameters defining the energy function. The randomness in the data is akin to what is called ``quenched disorder'' in the physics literature.\footnote{In statistical physics, the term ``quenched'' refers to coefficients randomly picked at the preparation of the system and kept constant during its evolution.} Training is routinely performed by the Stochastic Gradient Descent (SGD), which consists in starting from random initial conditions and then letting the weights evolve dynamically towards configurations corresponding to low loss values. This process is, in fact, similar to what is called ``a quench" in physics. The quenching protocol corresponds to a sudden decrease of the thermal noise, usually done by lowering the temperature of the thermal bath, for a system which is initially prepared in equilibrium at very high temperature. The study of the dynamics induced by quenches has been one of the most important topics of out-of-equilibrium physics of the last decades~\citep{BiroliLesHouches}. The main model considered in the literature is based on stochastic Langevin equations, reminiscent of SGD and corresponding to an evolution governed by gradient descent plus random noise. Since the initial temperature is very high, the initial conditions for the dynamics are random, featureless and uncorrelated with the quenched disorder if present, again in strong analogy with DNNs. 
Disordered systems are known to display glassy dynamics after a quench, which means that the system gets stuck for long times in local minima~\cite{BiroliLesHouches, reviewBCKM, reviewBB, cugliandololeshouches}. Given the similarity between the training of DNNs and quenching of disordered systems, it may seem surprising that meaningful local minima with perfect accuracy on the training set are found~\citep{zhang2016understanding}.

In the current literature, several explanations are proposed to explain this paradox. Two quite different points of view emerge from it. One is that even though the loss function displays a very large number of local minima with different loss values, the dynamics during the training process allows the system to decrease the loss without barrier crossing and to converge towards quite low local minima that allow good generalization. In other words, the loss landscape is {\it very rough}, however, this doesn't damage the performance of the system. In this direction,~\cite{choromanska2015loss} proposed an analogy with mean-field glassy systems. In such systems, it was shown by theoretical physics methods~\cite{cuku}, backed up by rigorous results~\cite{eqBAetal}, that dynamics corresponding to gradient descent or stochastic versions of it tend without barrier crossing to the widest and the highest minima, despite the existence of deeper local and global minima. A complementary point of view, proposed in~\cite{baldassi2016unreasonable}, is that there exist rare and wide minima which have large basins of attraction and are reached without any substantial barrier crossing by the training dynamics.

Another quite different point of view is that deep neural networks work in a regime in which there are actually no spurious local minima that can trap the system during the training process. Several rigorous and numerical works, including but not limited to~\cite{freeman2016topology, hoffer2017train, soudry2016no}, suggest that the loss function, despite being non-convex, is characterized by a connected level set as long as one considers loss values above the global minimum. From this perspective, the dynamical evolution induced by the stochastic gradient descent corresponds to falling down in the loss landscape without barrier crossing. In this case, it is the absence of bad local minima, and consequently, the absence of roughness and glassy dynamics, that solves the previous paradox.

Beyond the above two seemingly contradictory pictures on the structure of the loss landscape, there is also a rich literature discussing the path the dynamical process takes during the training process. For instance,~\citet{dauphin2014identifying} claims that it is the existence of numerous saddle points that lie on the dynamical paths that present itself as a form of an obstacle to find deeper local minimum. Several other works, including~\citet{lee2016gradient}, claim that gradient-based training avoids such obstacles even if they do exist. And finally, \citet{lipton2016stuck} demonstrates how the weights travel large distances through the flat basins by looking at the principle components of the evolution of the weights.

Establishing conclusively these scenarios in realistic cases is a challenge. Exact calculations of the statistical properties of critical points are hampered by the increased computational complexity of over-parametrized models and the possible degeneracy of critical points. Some guidance is provided by empirical results. In fact, simulations in~\citet{sagun2014explorations} demonstrate that different dynamical processes on the loss landscape can indeed perform similarly regardless of the effect of the noise of SGD, thus suggesting that barrier crossing indeed does not take place. The works~\citet{keskar2016large} and~\citet{jastrzkebski2017three} claim that by tuning the hyper-parameters of the system one can locate local minima with different qualities, thus providing indications of the roughness of the loss landscape. The results of~\citet{chaudhari2016entropy} demonstrate that wider and possibly rarer basins can be found by averaging out the values of several parallel optimizers. 

At the moment, it is still not clear what approach provides a good answer. It could be actually that the correct one contains ingredients from all the perspectives cited above. In this work, we address this problem by taking advantage of knowledge gained in the field of glassy out-of-equilibrium systems in the last decades~\cite{reviewbray, BiroliLesHouches, reviewBCKM}. Our approach is twofold: (1) probing the training dynamics through the measurement of one and two-point correlation functions, as done in physics, we infer properties of the loss landscape in which the system is evolving, (2) comparing the results obtained for mean-field glasses to measurements performed for realistic DNNs we test the analogy between these systems. 

\textbf{Our Contribution:} The analysis is performed for several different architectures, see Sec.~\ref{sec:models}, varying from specific toy models to ResNets~\citep{he2016deep} which are evaluated on popular datasets such as MNIST and CIFAR. We decided to focus both on a simple architecture and on more competitive ones. The former is close to a model where, for a large-enough hidden layer, there is a proof of the non-existence of bad local minima~\citep{freeman2016topology}, and the latter are a relatively more realistic one with relevant performances on the given task. The dynamical behavior we found is similar for all cases: After an initial exploration of high-loss configurations, the system starts its descent in the ``loss landscape'', and displays a particular kind of glassy dynamics, called \emph{aging}, see Sec.~\ref{sec:glassy}. Our results suggest that the slowness of the dynamics in this stage is not related to the crossing of large barriers but instead to the emergence of an increasingly large number of flat directions~\citep{sagun2017empirical}. At long times, a stationary regime where aging is interrupted and the system becomes almost stationary sets in. We present evidences that this dynamical regime corresponds to diffusion, not necessarily isotropic (as suggested by~\cite{jastrzkebski2017three}), at or close to the bottom of the loss landscape. We compare these behaviors to the ones of the $p$-spin spherical model, which is one of the most studied mean-field glass models. 
We find that although the first regimes share similarities with the dynamics of mean-field glasses after a quench, the final regime does not. This suggests a qualitative different geometrical characterization of the bottom of the loss landscape and, accordingly, of the dynamics within it.

\section{Basic facts on glassy dynamics}
\label{sec:glassy}

\begin{figure*}
    \subfigure[Energy of the $p$-spin model.]{%
    \label{fig:pspin-energy}%
    \includegraphics[width=\columnwidth]{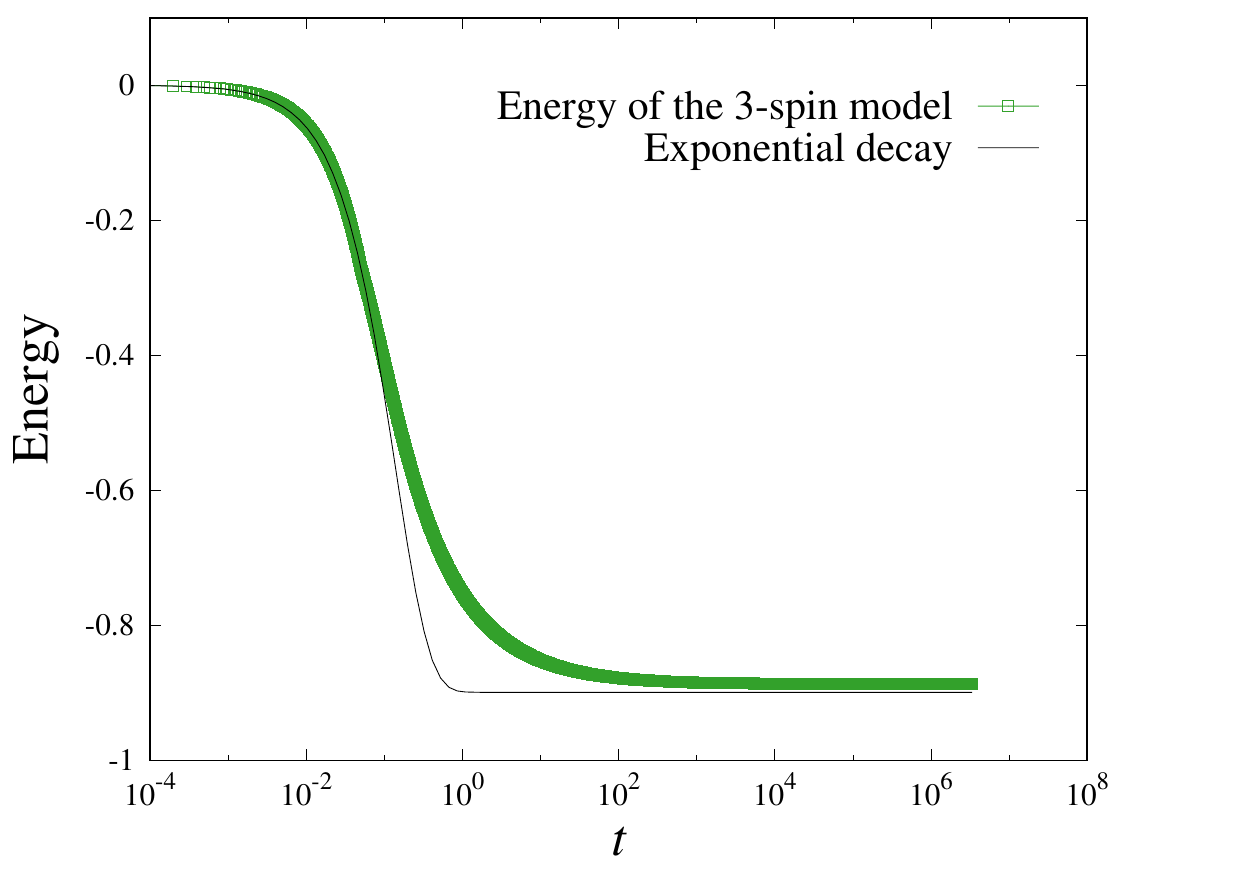}}%
    \qquad
    \subfigure[Mean square displacement of the $p$-spin model.]{%
    \label{fig:pspin-msd}%
    \includegraphics[width=\columnwidth]{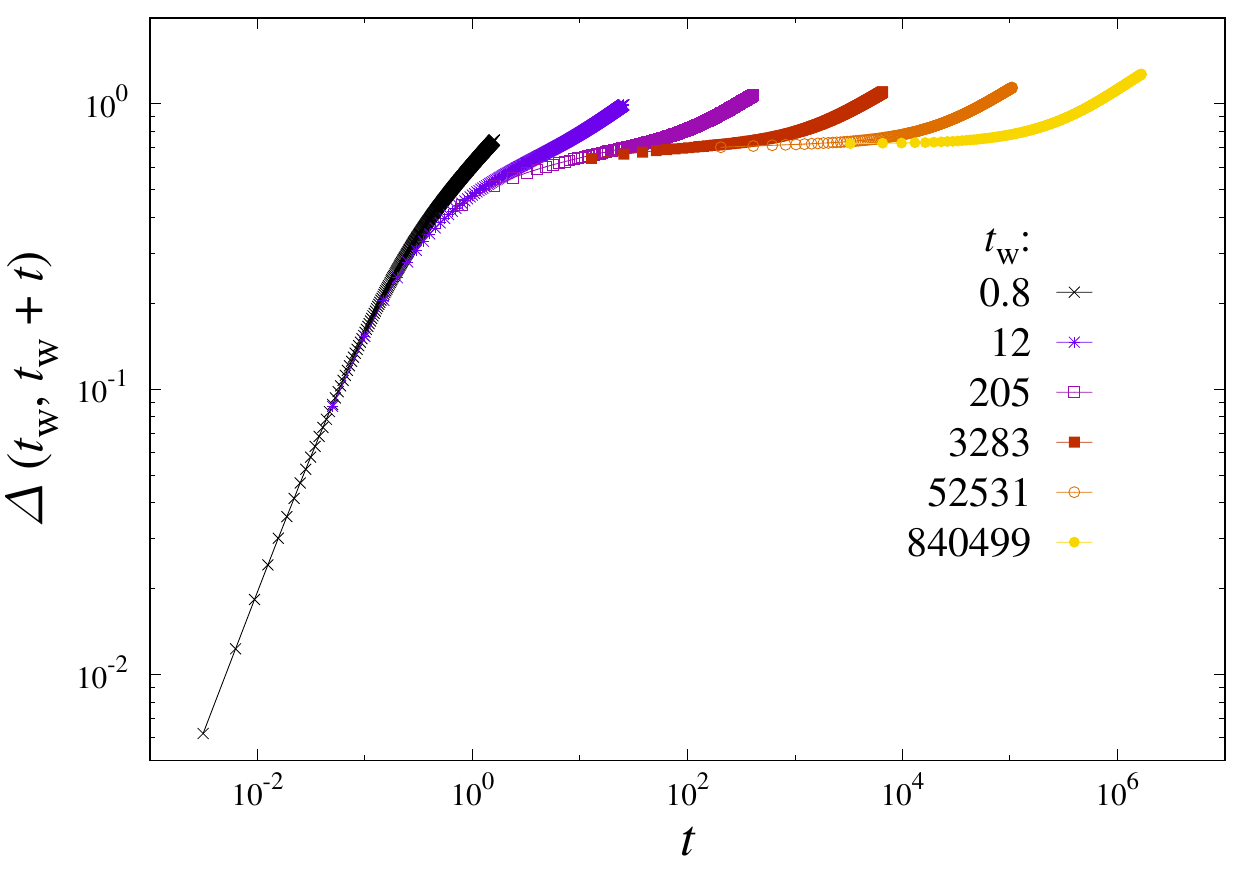}}%
    \caption{Energy,~\ref{fig:pspin-energy}, and the mean square displacement,~\ref{fig:pspin-msd}, of the $p$-spin model as a function of time in logarithmic scale after a sudden quench from a temperature $T_i=\infty$ to a temperature $T_f=0.5$, for $p=3$. In Figure~\ref{fig:pspin-energy}, we also show an exponential decay, for comparison. In Figure~\ref{fig:pspin-msd}, the mean-square displacement is displayed for several $\tw$, increasing from left to right.}
\end{figure*}

Two main observables have been identified as central to characterize the slow dynamics of physical systems. The first one is the energy as a function of time. When a system is quenched from high to low temperature the energy decreases and slowly approaches an asymptotic value. The functional dependence can be a power law of time, as in the Ising model~\cite{reviewbray}, or even a power of the logarithm of time as in several disordered systems, in particular glasses~\cite{reviewBB}. This dependence is called ``slow'' by comparison with an exponential relaxation which is typical of high-temperature phases\footnote{The existence of conserved quantities can produce a power-law dependence even in high-temperature phases.}. In Figure~\ref{fig:pspin-energy} we show the characteristic behavior of the energy as a function of time for a quench from high to low temperatures in the $p$-spin spherical model, which was highlighted in the context of DNNs through an analogy in~\cite{choromanska2015loss} and through phenomenological comparison in~\cite{sagun2014explorations}. The degrees of freedom of the $p$-spin model are $\sigma_i$, the $N$ components of a vector belonging to the $N$-dimensional sphere of radius $\sqrt N$. Its energy reads for $p=3$:
\begin{equation}
  \label{eq:3spin}
  E=-\sum_{\langle i_1,i_2,i_3   \rangle}J_{i_1,i_2,i_3}\sigma_{i_1}\sigma_{i_2}\sigma_{i_3}
\end{equation}
where the sum runs over all the possible $3$-tuples and $J_{i_1,i_2,i_3}$ are i.i.d. Gaussian random variables with zero mean and variance $3/N^2$. The dynamical evolution is governed by the stochastic Langevin equation. This model has a dynamical transition at a temperature $T_d\simeq 0.612$, see~\cite{cavagnaSGpedestrians} for a review. The plot in Figure~\ref{fig:pspin-energy} corresponds to a quench from $T_i=\infty$ to $T_f=0.5$; it is obtained by integrating numerically the Cugliandolo-Kurchan equations~\cite{cuku,eqBAetal}.

The second observable used to investigate out-of-equilibrium dynamics is the two-time correlation function. Its precise definition depends on the system at hand. For instance, in the case of the 3-spin model a possible choice is the mean-square displacement between $\tw$ and $\tw+t$:
\begin{equation}
  \label{eq:corr}
  \corrglass(\tw,\tw+t)=\frac 1 N \sum_{i=1}^N \left(\sigma_i(\tw)-\sigma_i(\tw+t)\right)^2  
\end{equation}
The correlation function is a measure of how much the configuration of the system at time $\tw+t$ decorrelates from the one at time $\tw$. The two times are chosen in order to explicitly probe the out-of-equilibrium nature of the dynamics: $\tw$ is the time lapse after the quench, $t$ is the difference between the two times at which system configurations are compared. When the system is out-of-equilibrium, in particular after the quench, $\corrglass(\tw,\tw+t)$ explicitly depends on both $\tw$ and $t$, whereas when equilibrium is reached the system becomes stationary and $\corrglass(\tw,\tw+t)$ only depends on $t$. When quenched at low temperature many disordered systems show the phenomenon of \emph{aging}, which means that the time-scale controlling the $t$-dependence is a function of $\tw$. In other words, the time it takes for the system to decorrelate depends on the age of the system.

In Figure~\ref{fig:pspin-msd}, we plot $\corrglass(\tw,\tw+t)$ for the 3-spin model as a function of $t$ and for different values of $\tw$. Focusing on the $t$-dependence, one can recognize the first time regime, which appears almost independent of $\tw$, in which the system appears stationary. This regime eventually ends at a time that increases with $\tw$. Then, the second regime which physically corresponds to aging emerges\footnote{The large-time limit of $\corrglass(\tw,\tw+t)$ is equal to two, as it should for diffusion on a sphere, where displacements are bounded. In Figure~\ref{fig:pspin-msd}, this limiting behavior is not seen because the simulations have been stopped early.}. Here, the longer is $t_w$ the longer it takes for the system to diffuse, i.e. for the mean-square displacement to escape from the plateau value. The height of the plateau is called Edwards-Anderson parameter in the physical literature and quantifies how much the system is frozen into a local minimum~\citep{cavagnaSGpedestrians}. 

Slow dynamics and aging are distinctive features of \emph{any} glassy system. Particularly, in the $p$-spin spherical model, and in other models of glasses, the slow dynamics observed after a quench\footnote{This dynamical regime corresponds to large time-scales that do not diverge with $N$. There is a second regime of time-scales, that diverge exponentially with the number of degrees of freedom~\cite{montanarisemerjian,benarousj}, in which barrier crossing does take place. In practice, except for small systems~\cite{BJREM}, this second regime cannot be accessed numerically since the corresponding time-scales are too big.} is \textit{not} due to barrier crossing but to the emergence of almost flat directions~\cite{cavagnaSGpedestrians}. As explained in~\cite{kurchanlaloux}, this phenomenon is due to the peculiarity of gradient descent in very high-dimensions; in this case the system is always confined at the border of the basins of attraction, and the Hessian at long times contains a decreasing number of negative eigenvalues, thus leading to an increasingly slow dynamics.

\section{Models and Results}\label{sec:models}
\label{sec:results}

We present our core results in two parts: time dependence of the loss function (Sec.~\ref{sec:loss}), and identifying different regimes through the two-point correlation function (Sec.~\ref{sec:msd}). We start by describing the models used for evaluation:
 \footnote{We did not remark any significant difference in the presence of explicit regularization, so we present the results where no regularization is used.}

\textbf{A - Toy Model:} The network contains only 1 hidden layer with $10^4$ hidden nodes. The non-linear function on the hidden layer is ReLU. The output layer is filtered through a sigmoid. The loss function is a mean square error. The total number of weights is around $3\times10^8$.

\textbf{B - Fully Connected:} A simple network with three fully connected layers, of sizes 100, 100 and 10, respectively. The non-linear functions are ReLUs, and the loss function is the negative log-likelihood of soft-max outputs. The total number of weights is about $9\times10^4$.

\textbf{C - Small Net:} A simple convolutional network with two conv-layers that has 10 and 20 filters in the first and second layer, respectively. It is followed by two fully-connected layers of sizes 100 and 10. The non-linear functions in the hidden layers are ReLUs, and the loss function is the negative log-likelihood of soft-max outputs. The total number of weights is around $6\times10^4$.



\textbf{D - ResNet18:} The final model is a ResNet with 18 hidden layers. The total number of weights is around $2\times10^7$.

We have chosen networks with various levels of complexity. All networks are initialized in the standard procedures of the PyTorch library (version 0.3.0). The toy model is inspired by the one introduced in~\cite{freeman2016topology} which is shown \textit{not} to have any barriers if the hidden layer is large enough. The training is carried out by SGD that takes a single learning rate that remains unchanged until the end of the computation. The training process runs for a fixed given number of iterations which is deemed to be `long enough' for all practical purposes. For most cases, this means that training kept running long after the perfect accuracy was reached on the training set. All the networks have been trained on multiple datasets: MNIST, CIFAR-10, CIFAR-100, and multiple sets of parameters.

\subsection{The Loss Function \label{sec:loss}} 

\begin{figure*}[ht]
    \subfigure[Toy Model on CIFAR-10 $m=10^4$, $B=100$, $\alpha=0.1$.]{%
    \label{fig:loss-A}%
    \includegraphics[width=\columnwidth]{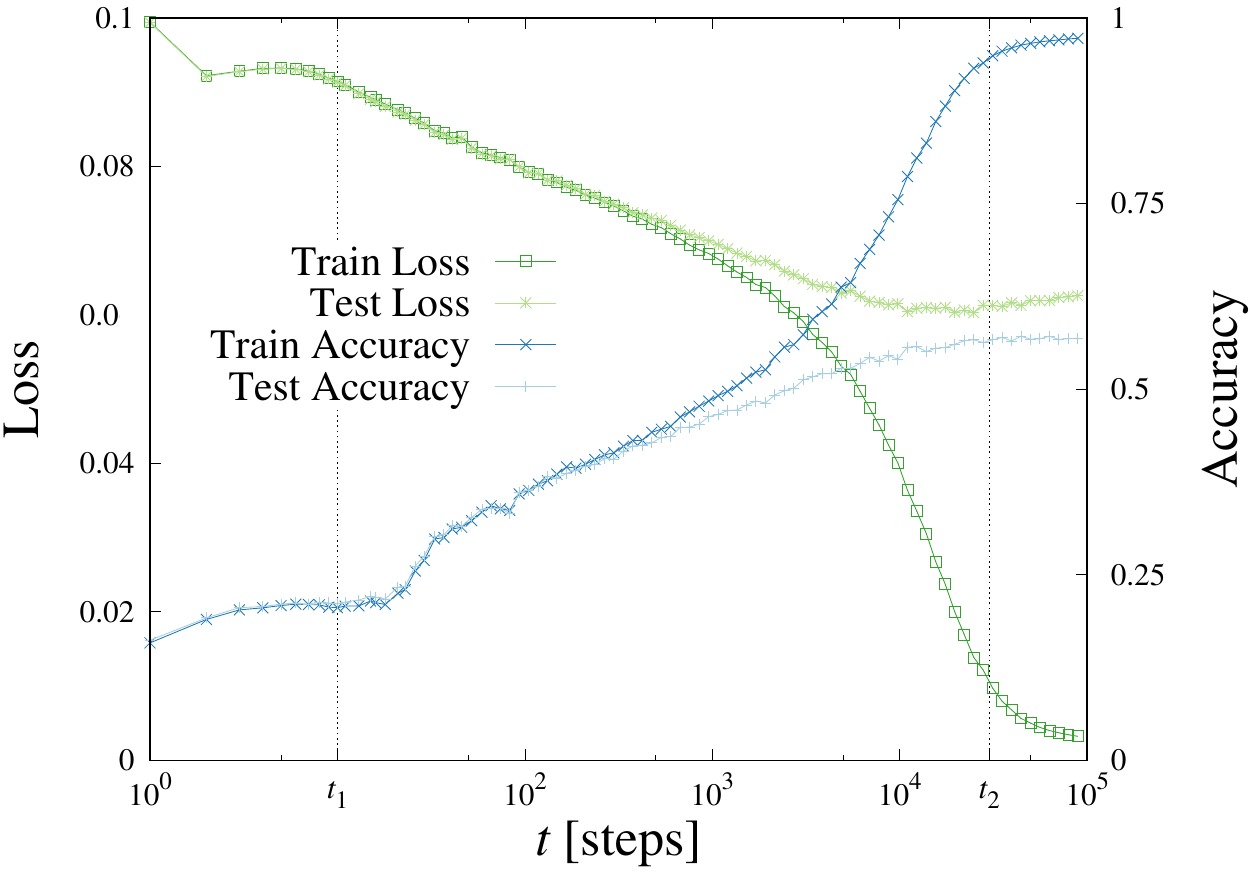}}%
    \qquad
    \subfigure[Fully Connected on MNIST, $B=128$, $\alpha=0.01$.]{%
    \label{fig:loss-B}%
    \includegraphics[width=\columnwidth]{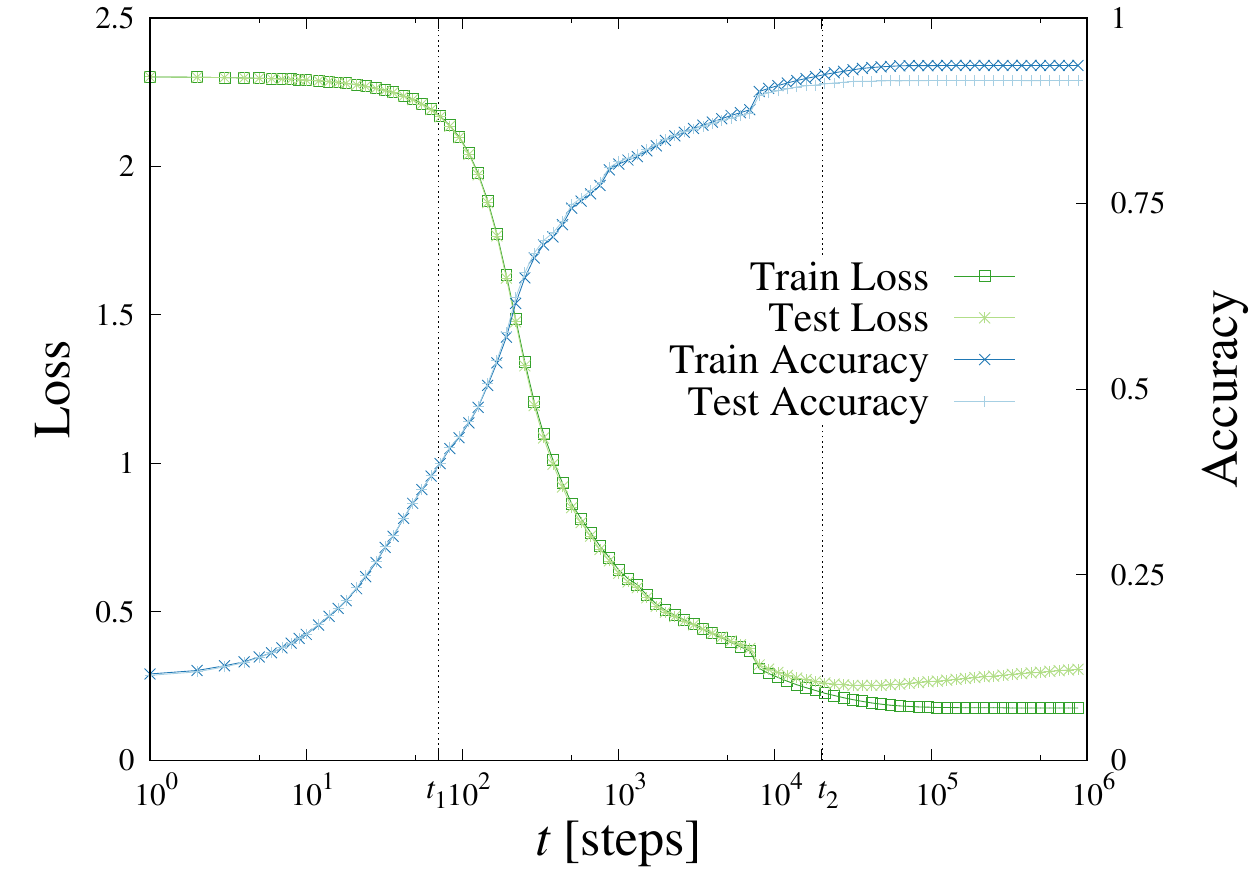}}
    \qquad
    \subfigure[Small Net on CIFAR-10, $B=100$, $\alpha=0.01$.]{%
    \label{fig:loss-C}%
    \includegraphics[width=\columnwidth]{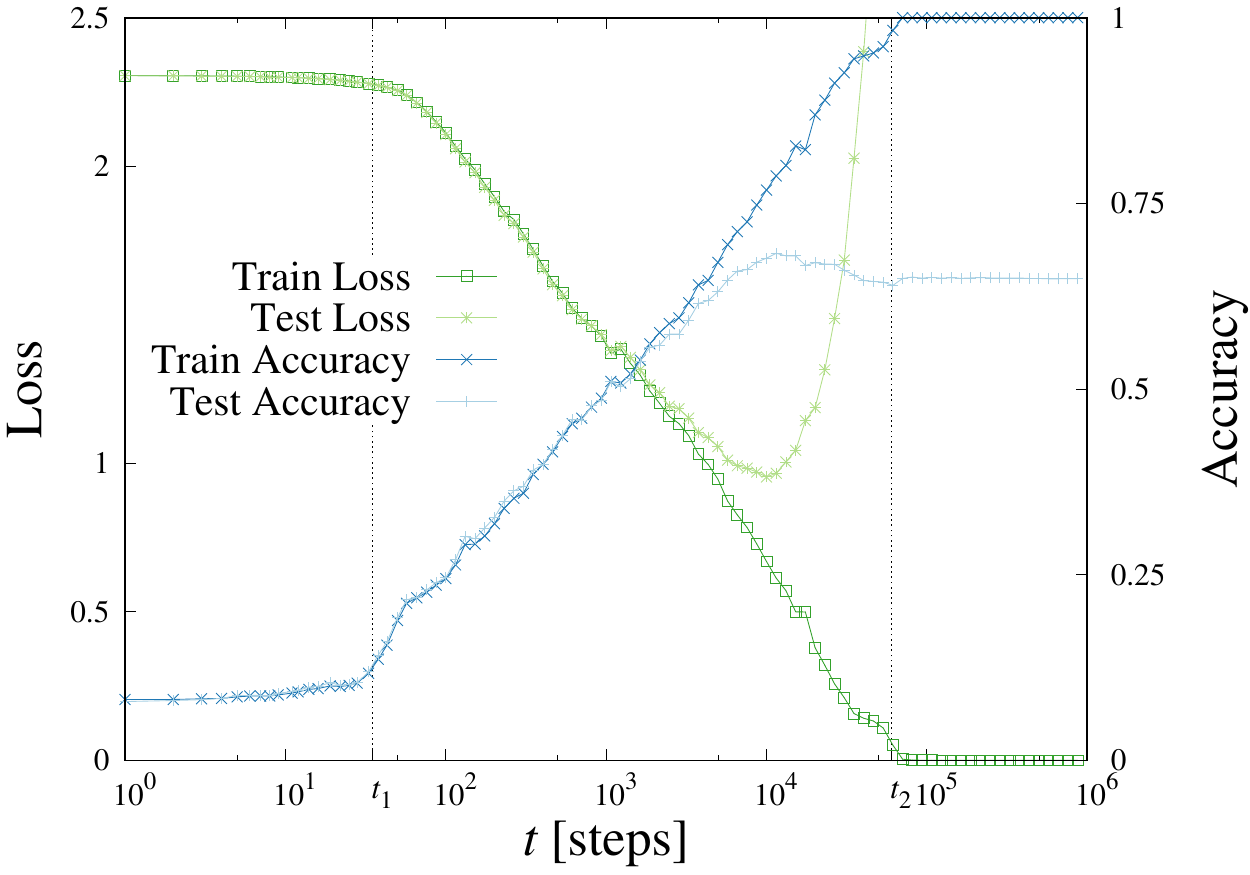}}%
    \qquad
    \subfigure[ResNet-18 on CIFAR-100, $B=64$, $\alpha=0.01$.]{%
    \label{fig:loss-D}%
    \includegraphics[width=\columnwidth]{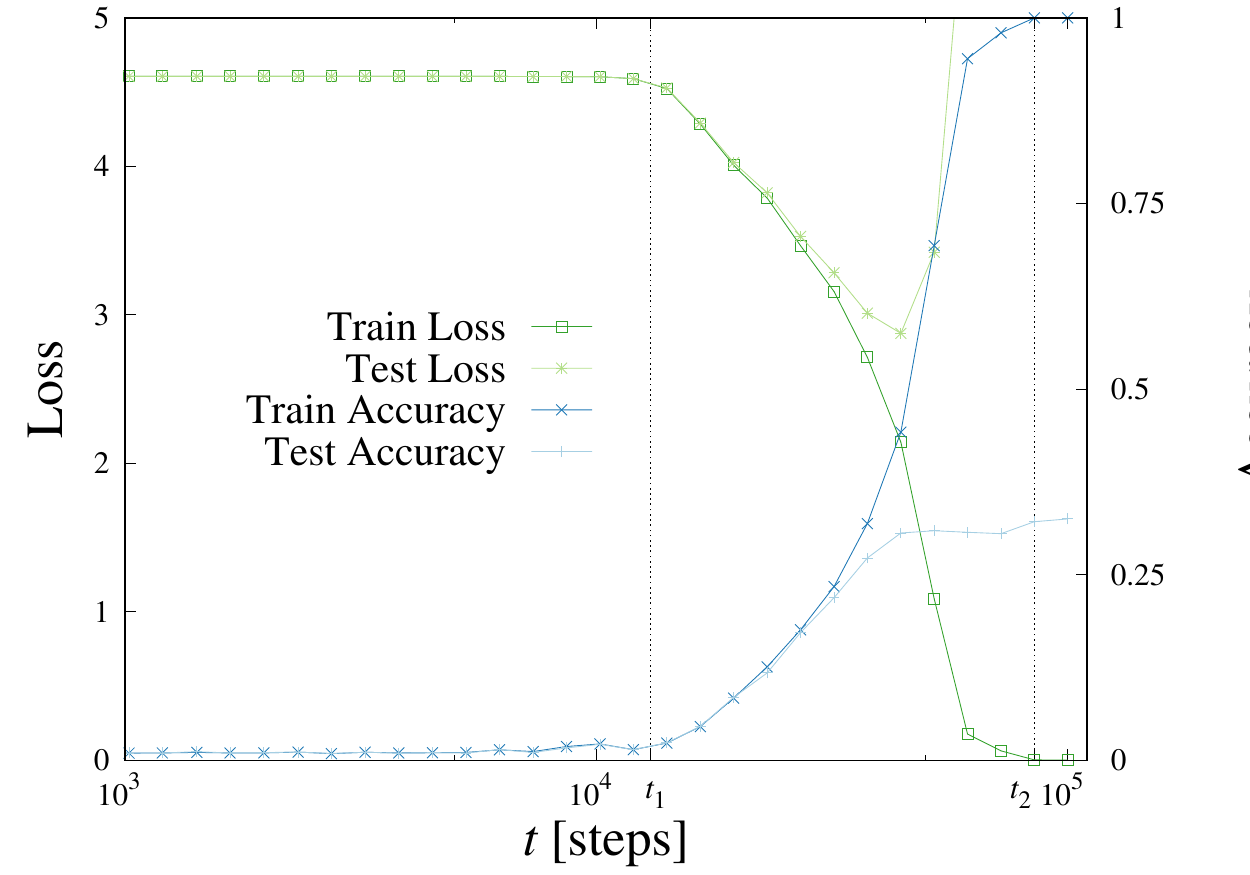}}%
    \caption{Train/test loss and accuracy as a function of $\log(t)$. The batch size $B$ and learning rate $\alpha$ are specified under each plot. Note that in~\ref{fig:loss-A} it is more difficult to pin-point the values of $t_1$ and $t_2$ since the crossover is not as sharp as in the other cases.}
\end{figure*}

We first focus on the time-dependence of the loss function over the training, and we compare it to the one of the energy in glassy systems. For the sake of completeness, we also show the accuracy. We plot the loss values as a function of the logarithm of time, measured in units of iterations so that the unit time step corresponds to a single update of the weights. This choice is different from the wall time or number of epochs which is often used. Although less common in machine learning, the logarithmic scale highlights the slow dynamics and the time dependence\footnote{A positive side effect of a logarithmic representation is that the measurements can be exponentially spaced. As a consequence, the numerical overhead of the measurements goes to zero as the simulation time increases. Since the relevant time scales are logarithmic, this implies no loss of information.}. The results obtained for the four networks described above are shown in Figures~\ref{fig:loss-A},~\ref{fig:loss-B},~\ref{fig:loss-C},~\ref{fig:loss-D}. There are several features worth noticing. We can remark three regimes. The first one goes from the beginning of the training up to a time $t_1$, where the loss and accuracy stay roughly constant. At $t=t_1$ the loss starts decreasing roughly linearly in $\log(t)$, and concomitantly the accuracy increases in a similar way. This second regime persists until a time $t_2$, at which the train loss approaches zero. In the final regime beyond $t_2$ the speed of decay sharply decreases. The cross-over times $t_1$ and $t_2$ are indicated in Figures~\ref{fig:loss-A},~\ref{fig:loss-B},~\ref{fig:loss-C},~\ref{fig:loss-D}.
In Sec. \ref{sec:msd} we show that $t_1$ and $t_2$ can also be identified through the evolution of the mean-square displacement.

This behavior is similar to the ones found in disordered systems, see e.g. Figure~\ref{fig:pspin-energy}. There are however two main differences. First, in several cases the decrease in the second part is actually slower for the DNNs compared to the power-law of the $p$-spin model\footnote{The power law decrease of the energy was established in~\cite{cuku} and is well verified numerically.}. Second, and more importantly, the loss reaches asymptotically (i.e. after $t_2$) its lowest possible value. This is not the case in the $p$-spin model in which instead the energy converges asymptotically to one of the highest and widest minima~\cite{cuku,cavagnaSGpedestrians}. Actually, a $p$-spin model with a number of degrees of freedom comparable to the number $M$ of weights that are used in deep learning (in our examples $M=10^4-10^7$) would take an exponentially long time to go beyond the highest and widest minima and reach the bottom of the landscape~\cite{cavagnaSGpedestrians,reviewBB}. This is a first indication that the dynamics involved in the training of deep neural networks, although slow, does not correspond to the crossing of large barriers, which would instead lead to much longer time-scales.


In summary, the reason for the slowing down of the dynamics during training is apparently not due to barrier crossing but instead likely related to an increasingly large amount of flat directions that become available to the system during its descent in the loss landscape, as found numerically in~\citep{lecun1998efficient, sagun2017empirical}. This is actually similar in the $p$-spin spherical model to the first dynamical regime of aging dynamics that follows a quench. However, in this case the system does not reach the lowest possible values of the loss, as it happens to loss functions during training, but remains trapped in higher and wide local minima. 


\subsection{Further evidence: Two-time correlation functions \label{sec:msd}} 

In this section, we focus on the two-time mean-square displacement $\Delta(\tw,\tw+t)$ of the weights and we compare it to the one found for disordered systems after a quench. Its definition reads:
\begin{equation}\label{eq:msd}
    \Delta(\tw,\tw+t)=\frac 1 M \sum_{i=1}^M \left(w_i(\tw)-w_i(\tw+t)\right)^2
\end{equation}
where the sum runs over all the weights $w_i$ of the network, and $M$ is their total number. 

\begin{figure}[ht]
    \centering
    \includegraphics[width=\columnwidth]{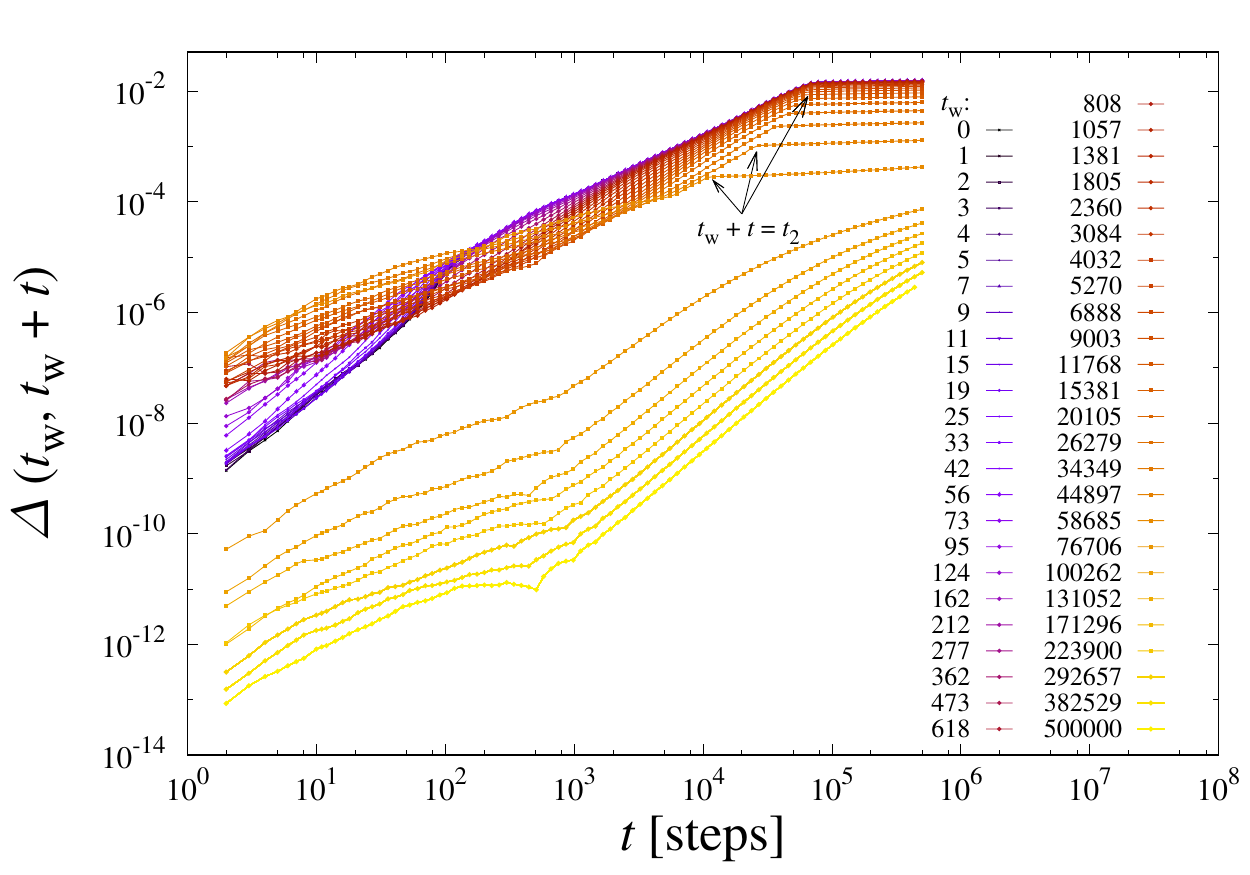}
    \caption{Two-time mean square displacement, $\Delta(\tw,\tw+t)$, defined in Equation~\ref{eq:msd}, for model C (Small Net). Every curve corresponds to a different waiting time $\tw$, indicated in the legend.}
    \label{fig:msd}
\end{figure}

The three regimes of the learning dynamics described in Sec.~\ref{sec:loss} are visible also through the behavior of the mean-square displacement. In Figure~\ref{fig:msd}, for $\tw<t_1$, $\Delta(\tw,\tw+t)$ collapses on a single curve. Once $t_1<\tw<t_2$ the mean-square displacement develops a clear dependence on $\tw$: the characteristic time increases with $\tw$, thus showing aging, and when $t>t_2-\tw$ it suddenly becomes flat. In the third regime, which corresponds to $t_w>t_2$, the characteristic time does not increase any longer with $\tw$. 

To a large extent, the training dynamics at large times can be explained in terms of diffusion in the weight space. A hallmark of a diffusing system is a motion purely driven by the noise $D$~\cite{crank1979mathematics}. We estimate the noise in SGD with the variance of the loss function's gradient\footnote{For reasons of numerical efficiency, for some models $D$ is calculated on a (sufficiently large) subset of the training set.}, which reads (details on the definition of the noise can be found in several resources, see, for example, \citet{li2015dynamics}):
\begin{equation}
    D = \frac{1}{|\text{train set}|} \sum_{s \in \text{train set}} \frac1M \left | \nabla \mathcal{L}_s - \nabla \mathcal{L} \right |^2
\end{equation}
where $\mathcal{L} = \frac{1}{|\text{train set}|} \sum_{s \in \text{train set}} \mathcal{L}_s$ is the empirical average and $\mathcal{L}_s$ is the loss of the $s$-th image in the train set. In a glassy system, the noise is constant through time if the temperature is fixed, whereas during the training $D$ varies, being a function of the network's weights. When comparing the results obtained at different $\tw$ we then normalize the mean-square displacement by $D(\tw)$, since larger $D(\tw)$ leads naturally to larger $\Delta(\tw,\tw+t)$, as illustrated by simple diffusion processes\footnote{The normalization by $D(\tw)$ is just an approximate way to take into account the variation of the noise with time; it works well if the variation is not too fast compared to $t$.}.

\begin{figure*}[ht]
    \subfigure[Toy Model on CIFAR-10, $B=100$, $\alpha=0.1$.]{%
    \label{fig:corr-rescaled-A}
    \includegraphics[width=\columnwidth]{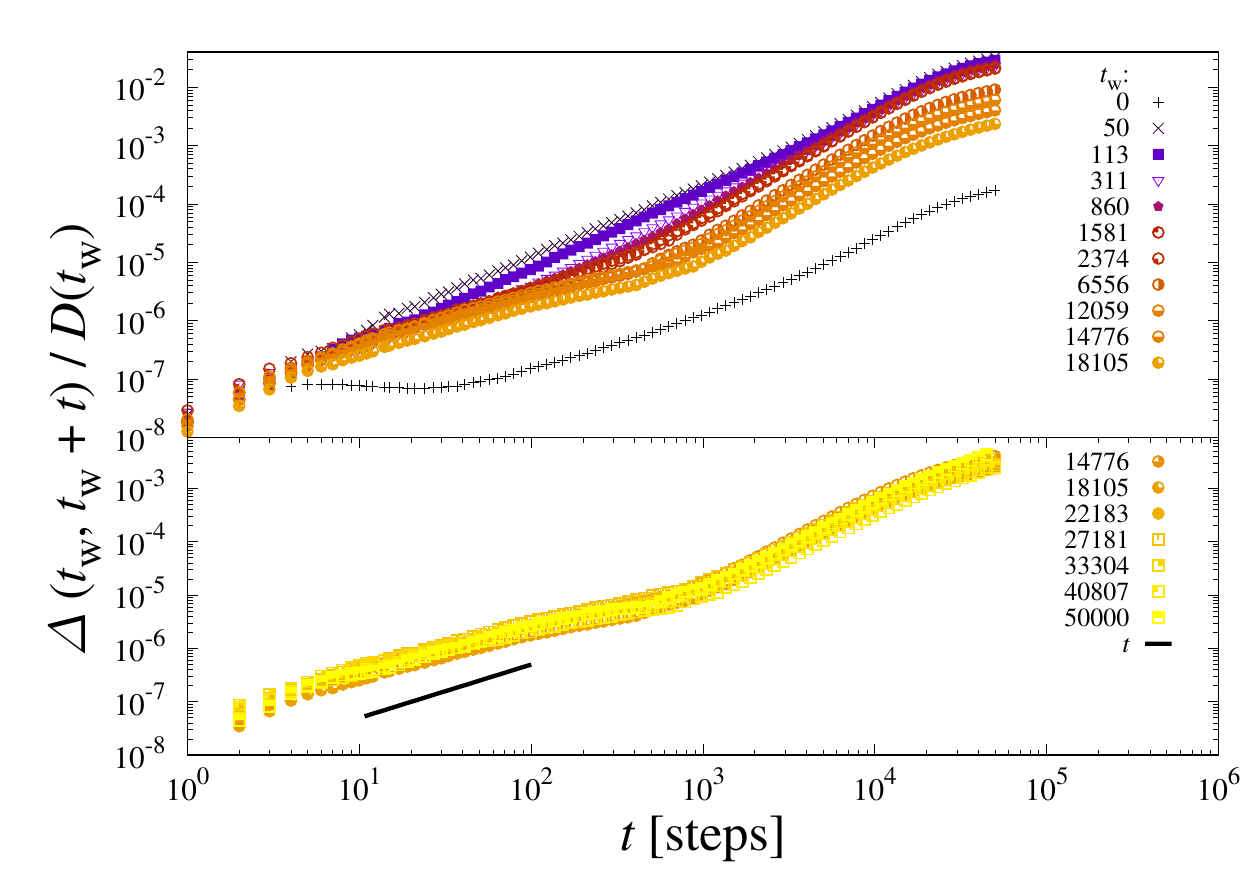}}
    \qquad
    \subfigure[Fully Connected on MNIST, $B=128$, $\alpha=0.01$.]{%
    \label{fig:corr-rescaled-B}
    \includegraphics[width=\columnwidth]{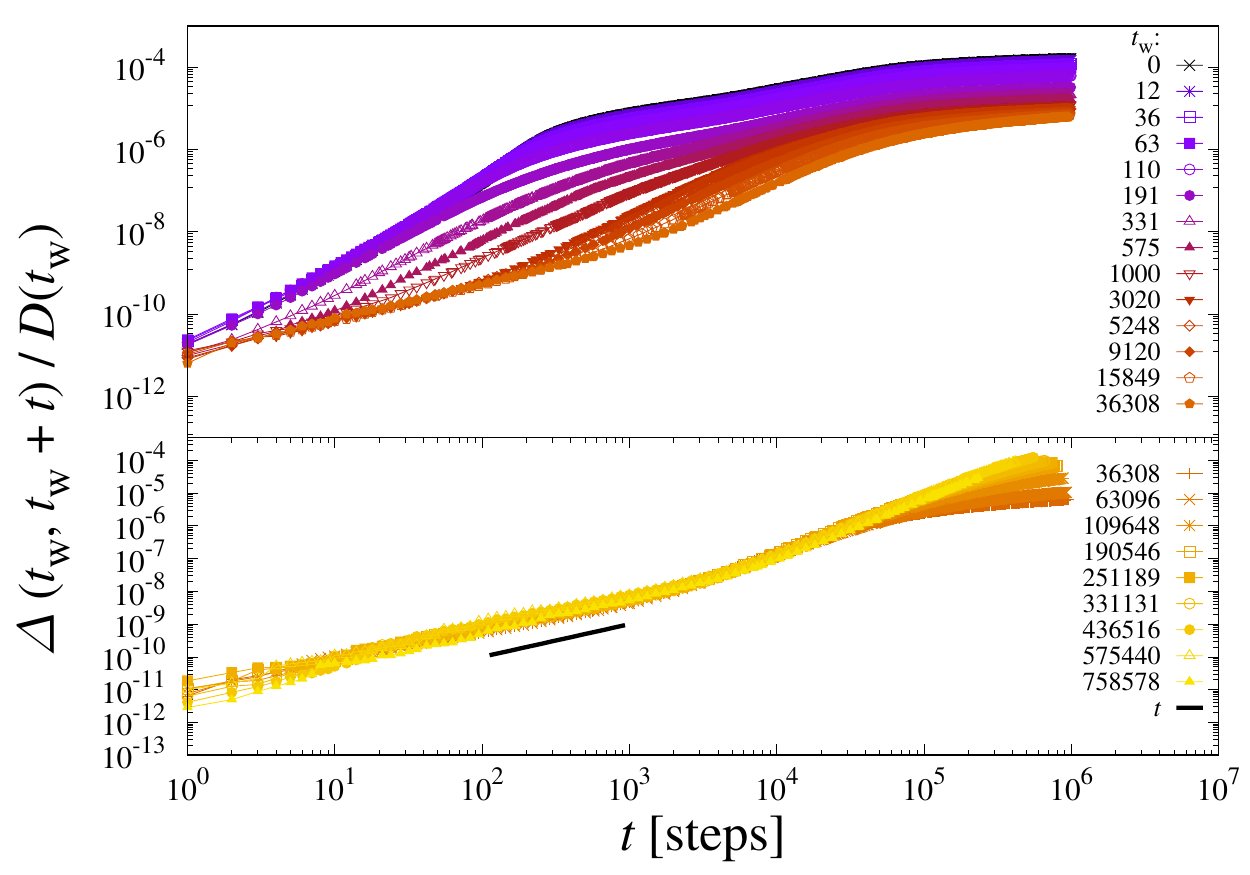}}
    \qquad
    \subfigure[Small Net on CIFAR-10, $B=100$, $\alpha=0.01$.]{%
    \label{fig:corr-rescaled-C}
    \includegraphics[width=\columnwidth]{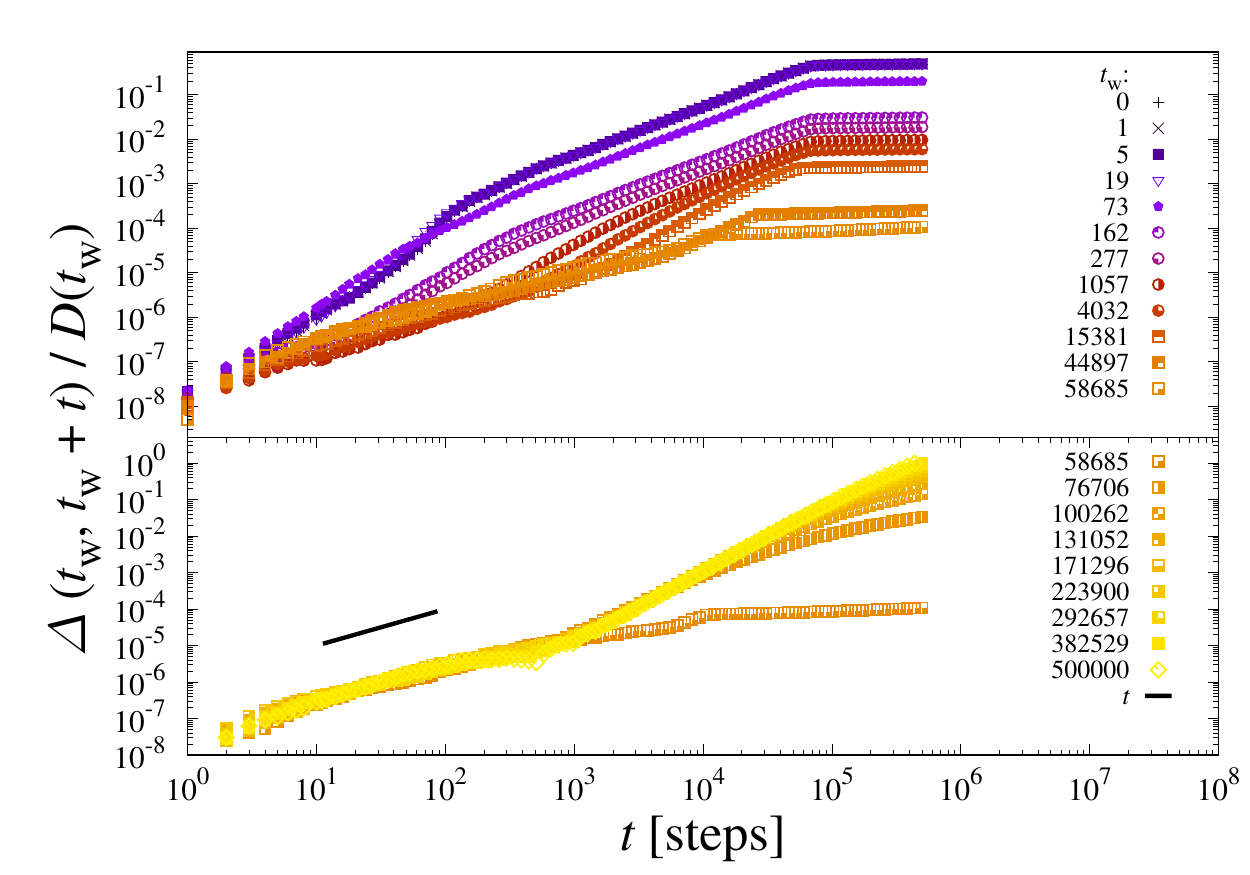}}
    \qquad 
    \subfigure[ResNet-18 on CIFAR-100. $B=64$, $\alpha=0.01$.]{%
    \label{fig:corr-rescaled-D}
    \includegraphics[width=\columnwidth]{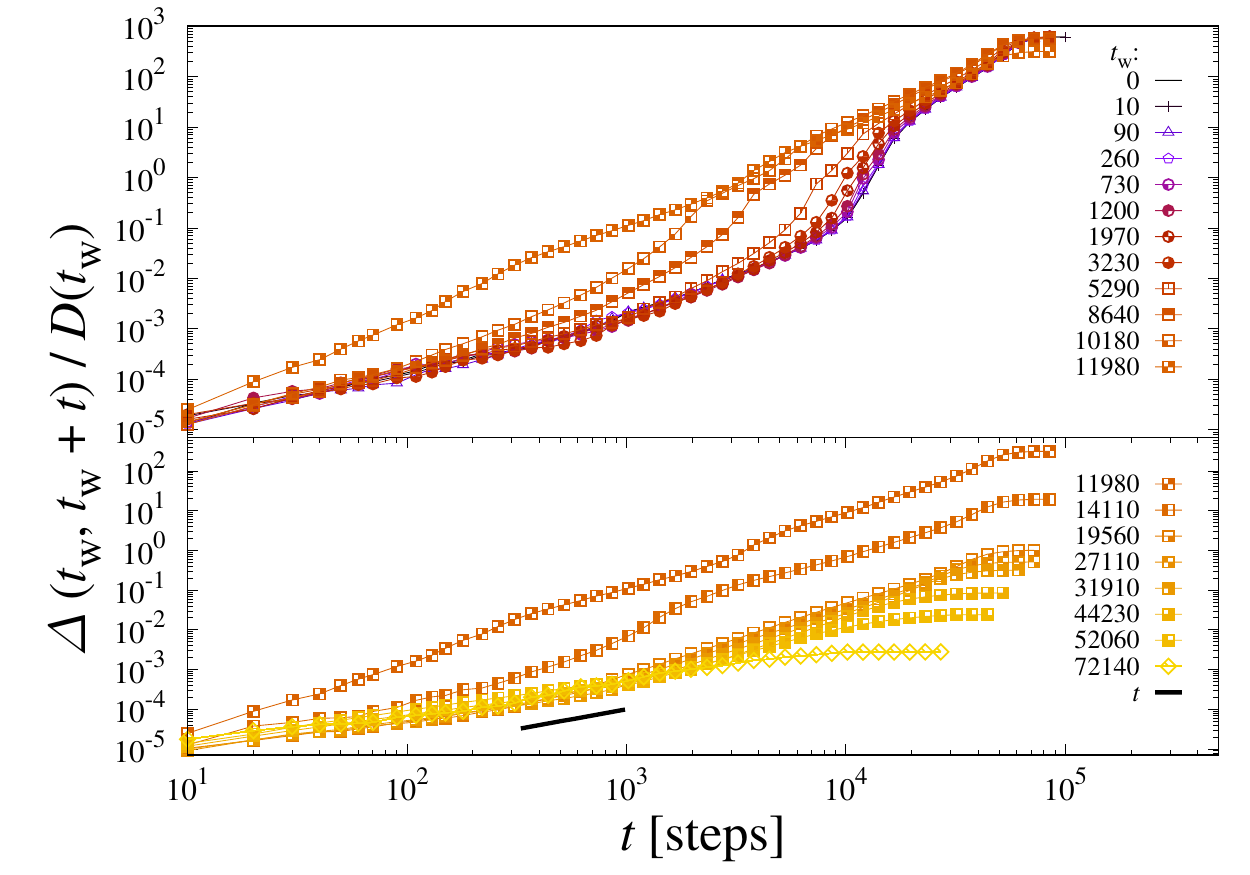}}
    \caption{Mean square displacements rescaled by the noise on the loss's gradient. Since the behavior of the curves differ in different phases, we show the smaller $\tw<t_2$ on the top set, and the larger $\tw>t_2$ on the lower set. For reference, some $\tw$ appear in both sets. The black segment on the bottom sets represents a slope $\sim t$.}
\end{figure*}

We present the mean square displacements in Figures~\ref{fig:corr-rescaled-A},~\ref{fig:corr-rescaled-B},~\ref{fig:corr-rescaled-C},~\ref{fig:corr-rescaled-D}\footnote{For model B and D we averaged over eight and two random initializations, respectively. This is done to iron out the fluctuations of the mean-square displacement. In principle, in order to see the collapse, this procedure should have been carried out for all experiments, but it was not required for models A and C.}. The main result that we find is that for $\tw<t_2$ there is a clear $\tw$ dependence, whereas at larger times the curves for different $\tw$ collapse together when scaled with $D(\tw)$. To stress this fact each of the plots has been split in two panels: the upper one shows the curves with $\tw<t_2$ and the lower one those with $\tw>t_2$
\footnote{Except Fig.~\ref{fig:corr-rescaled-D}, where we could not reach long-enough times, and a hybrid regime is represented.}. The collapse indicates that, except for the change in the strength of the noise $D$, the dynamics is reaching a stationary regime for $\tw>t_2$. In this regime, the loss function is almost equal to zero, thus indicating that the system is diffusing close to the bottom of the landscape.\footnote{Notes on further experiments: (1) LeNet on CIFAR10 with 77\% test accuracy presents collapse curves at least as good as Figure~\ref{fig:corr-rescaled-C}, and (2) Deeper ResNet \& WideResNet models on both CIFAR10 and CIFAR100 with better accuracies than model \textbf{D} give the correct diffusive slope in their mean square displacement curves but the collapse is not as good as in Figure~\ref{fig:corr-rescaled-D}. We believe that the key to resolve the collapse in models where number of parameters are much larger goes through a better calculation of the noise coefficient. As a matter of fact, $D$ changes with time, so rescaling $\Delta(\tw,\tw+t)$ by $D(\tw)$ can only work well for small $t$. This also explains why in Fig.~4 the expected slope $\Delta/D\sim t$ is only identified for not too large $t$. We will analyze these issues in detail in an upcoming work.}

Let us now compare this situations with the one of physical systems after a quench, in particular the $p$-spin spherical model for $p=3$. In both cases one finds somewhat similar regimes characterized by aging, and corresponding to the descent in the loss (or energy) landscape. The behavior at large times is instead different. In the training dynamics aging is interrupted, meaning that the system becomes stationary except for the change in the noise strength, whereas for the $p$-spin model aging persists even when the energy approaches its asymptotic value (on time-scales that do not diverge with the system size). Another difference is the shape of the mean-square displacement curves. During aging, in Figure~\ref{fig:pspin-msd}, the curves follow a master curve  for small $t$ no matter what is the value of $\tw$, instead for DNNs no collapse at short-times is present. For $\tw>t_2$ the shape of the mean-square displacements does not show any intermediate plateau\footnote{The shape of the mean-square displacements is different for different networks, possibly indicating that the manifolds corresponding to the bottom of the landscape have different geometric characterization.}, contrary to what found in Fig.~\ref{fig:pspin-msd}. The form of $\Delta(\tw,\tw+t)$ is instead the one characteristic of diffusion (the curves $\Delta/D$ would be straight lines in a log-log plot only if $D$ didn't depend on $\tw$).

Both the aging and the diffusive regimes are present and qualitatively similar in all the analyzed networks. The fact that a slow aging dynamics is also present in model A (Toy Model), that supposedly has no barriers (see  Sec.~\ref{sec:results}), strengthens the conclusion that the dynamics slows down because of the emergence of flat directions that ultimately lead to diffusion at or close to the bottom of the landscape. A deeper analysis of the finer properties of the diffusive regime will be studied in a forthcoming publication.



\section{Discussion}\label{sec:discussion}

\begin{figure*}[ht]
    \subfigure[Loss of the under-parametrized model.]{%
    \label{fig:loss-A-smallmcifar10}
    \includegraphics[width=\columnwidth]{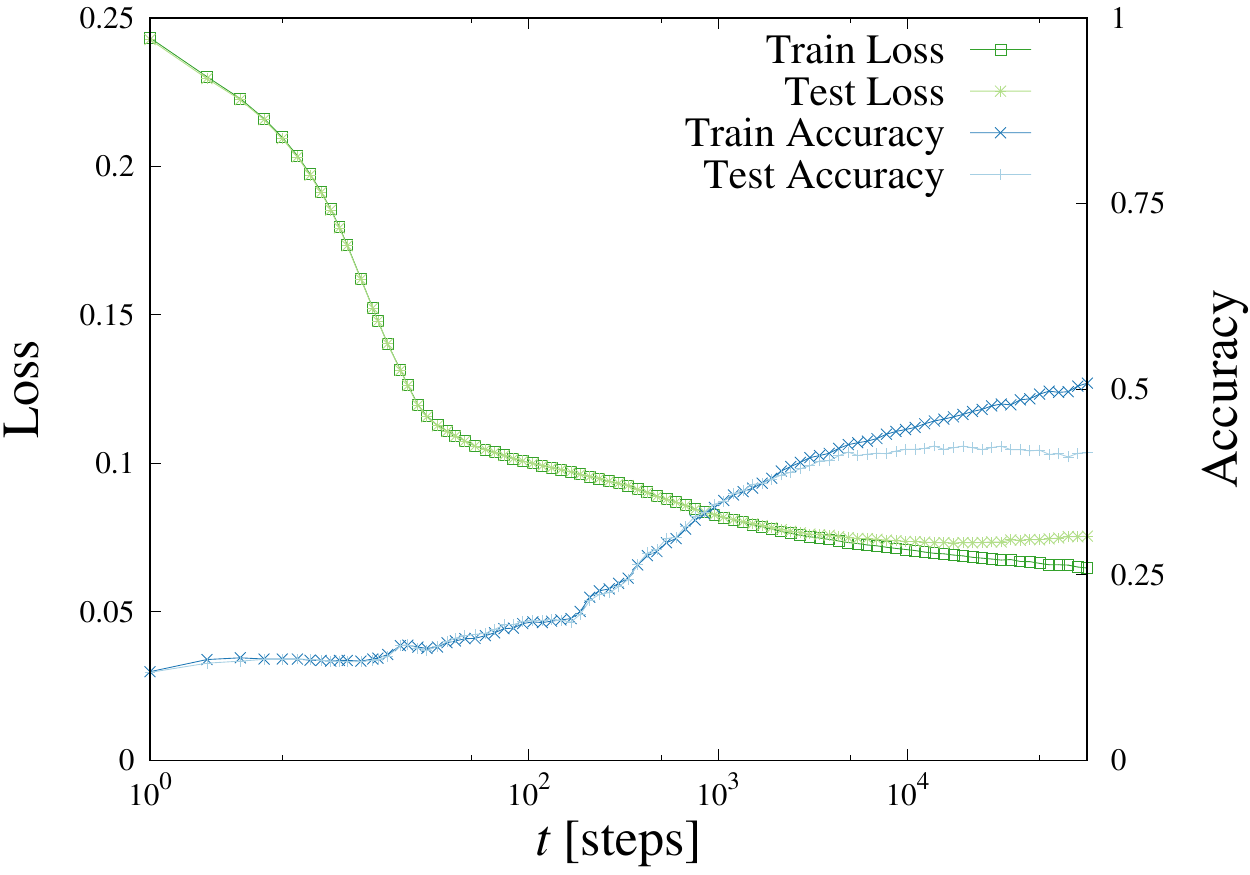}}
    \qquad
    \subfigure[Mean square displacement of the under-parametrized model.]{%
    \label{fig:corr-A-smallmcifar10}
    \includegraphics[width=\columnwidth]{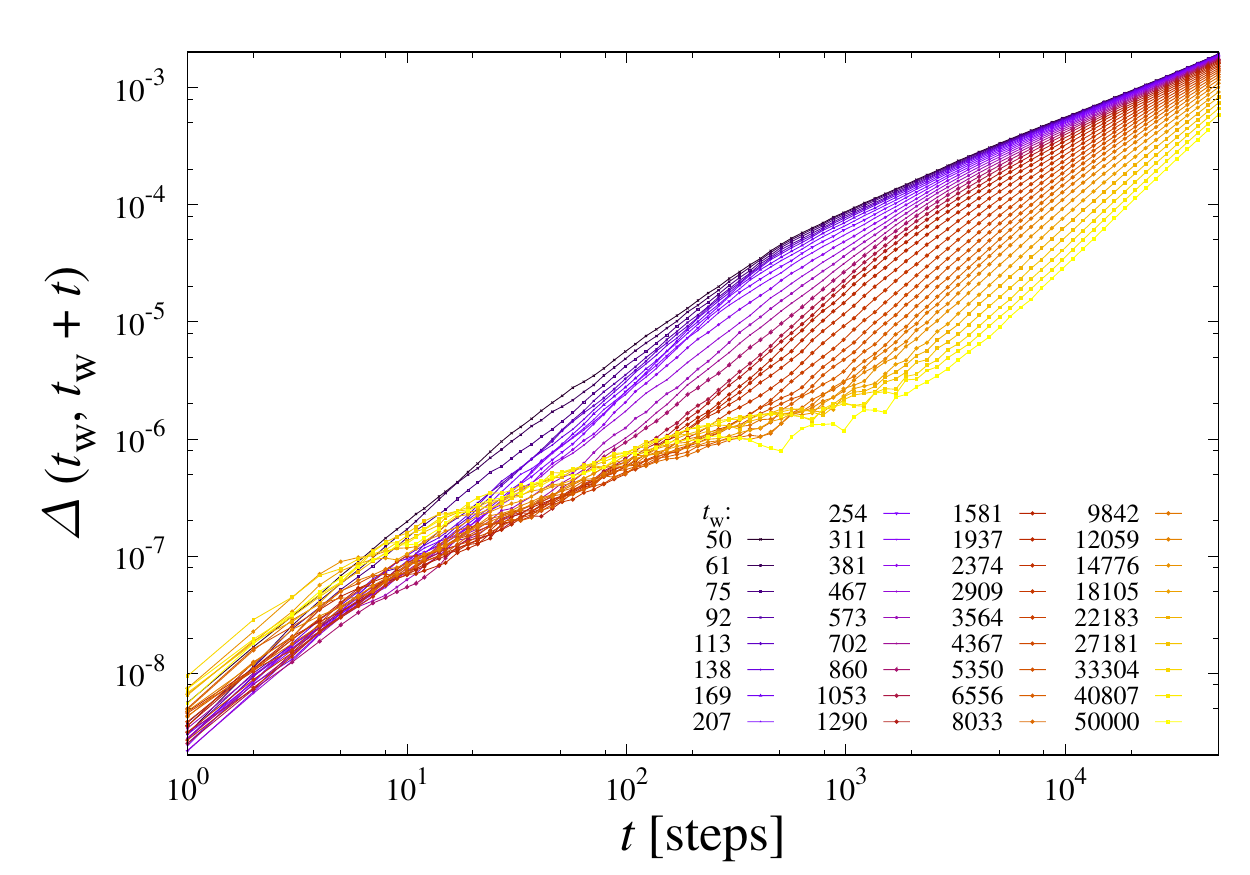}}
    \caption{On~\ref{fig:loss-A-smallmcifar10} train/test loss and accuracy as a function of $\log(t)$ in a modified version of model A (Toy Model) with only 10 hidden neurons on CIFAR-10. The batch size is $B=100$, and the learning rate is $\alpha=0.1$. On~\ref{fig:corr-A-smallmcifar10}, mean square displacement for the same model.}
\end{figure*}

In this work we have analyzed the training dynamics of DNNs by methods developed in physics for out-of-equilibrium disordered systems. We have studied the time dependence of the loss value and the mean-square displacements of the weights and compared them to their counterparts in physical systems, in particular the 3-spin spherical spin-glass. The analysis of the time-dependence of the loss function and the mean square displacement indicates that there are at least three time regimes in the training process: one corresponding to an initial exploration of the energy/loss landscape, followed by a decrease of the loss, in which the system displays aging dynamics, and a final regime in which the dynamics appears to be almost stationary and diffusive. Barrier crossing does not seem to play any role. The slowing down can be instead traced back to an increasingly large amount of flat directions that become available to the system during its descent in the loss landscape. 

The non-existence of such barrier crossings has been already proposed in the machine learning literature and some indirect evidences where obtained in numerical works. In~\cite{freeman2016topology}, it is shown that in certain networks one can connect two different solutions by a path in the weight space in such a way that the loss doesn't increase by much, and the amount of increase diminishes as the size of the network grows. In a related perspective on the loss surface,~\cite{sagun2016singularity} and~\cite{sagun2017empirical} demonstrate separate cases where the \textit{straight line} between two weight configurations at the bottom of the loss landscape evaluates to the same loss value, in other words there are no barriers between these two points. 

Overall, our study shows that there are interesting analogies between DNNs and glassy mean-field models but also important differences: in both cases slow evolution along almost flat directions is a key ingredient to understand the dynamics, however in DNNs the shape of $\Delta(\tw,\tw+t)$ at large $\tw$ combined with the fact that the system is able to reach the bottom of the landscape suggests that the statistical properties of the loss landscape are not the same even qualitatively. A possible reason for this difference is the over-parametrization of DNNs, which, pictorially, stretches the rough landscape and makes its dynamical exploration easier. Indeed, the dynamics of glassy systems was recently shown to be greatly accelerated by adding continuous parameters \cite{ninarello2017models}. As explained in~\cite{brito2018theory} this flattens the landscape and allows to reach very low energy states without jumping over barriers. 


In order to test this idea, we have reduced substantially the number of nodes for model A keeping the same dataset used for the previous figures. In this case the loss function does not reach zero, actually it seems to tend asymptotically to a higher value, see Figure~\ref{fig:loss-A-smallmcifar10}. Even more striking is the behavior of the mean-square displacement, which is now qualitatively similar to those of glassy systems, as shown in Figure~\ref{fig:loss-A-smallmcifar10}. One sees both a collapse at small values of $t$ for different values of $\tw$, possibly indicating the emergence of an Edwards-Anderson parameter and trapping in bad local minima, and a later $\tw$-dependent time increase just like in regular aging of disordered systems. 

On the basis of these results, we conjecture the existence of a phase transition between two regimes: (i) an easy phase corresponding to over-parametrized networks, in which bad local minima do not play any role, dynamics is governed by a massive amount of flat directions, and learning is achieved; (ii) a hard phase corresponding to under-parametrized networks, in which the landscape is rough, dynamics is glassy and the network does not learn well. Whether learning is possible in this case but it would take a huge amount of time to find the good minima is an interesting question. 

This scenario has tantalizing similarities with the one found in several combinatorial optimization problems in which easy, hard and impossible algorithmic phases have been found, see e.g.~\cite{monasson1999determining,mezard2002analytic,pnasmontanariksat,zdeborovareview,achlioptas2008algorithmic}. When degrees of freedom are continuous, the transition between these phases can be associated with the emergence of many flat directions in the energy landscape, a well-known example is the jamming transition of disordered solids \cite{Wyart05b,liu10}. A detailed investigation of this scenario for DNNs is ongoing and will be presented in a future publication.




\section*{Acknowledgements}
We thank Valentina Ros for useful conversations. We thank Utku Evci and U\u{g}ur G\"{u}ney for providing the initial version of the code that we used in our numerical simulations. This work was partially supported by the grant from the Simons Foundation ($\sharp$454935 Giulio Biroli, $\sharp$454953 Matthieu Wyart, $\sharp$454951 David Reichman). M.W. thanks the Swiss National Science Foundation for support under Grant No. 200021-165509.  M.B.-J. was partially supported through Grant No. FIS2015-65078-C2-1-P, jointly funded by MINECO (Spain) and FEDER (European Union). CC acknowledges support from the King’s Worldwide Partnership Fund.


\bibliography{main}
\bibliographystyle{icml2018}

\end{document}